\pdfoutput=1

\documentclass[11pt,dvipsnames]{article}

\usepackage[]{emnlp2021}

\usepackage{times}
\usepackage{latexsym}
\usepackage{graphicx}
\usepackage{booktabs}
\usepackage{enumitem}

\usepackage[T1]{fontenc}

\usepackage[utf8]{inputenc}

\usepackage{microtype}

\newcommand{\footremember}[2]{%
   \thanks{\xspace\xspace#2}
    \newcounter{#1}
    \setcounter{#1}{\value{footnote}}%
}

\graphicspath{{figs/}{../}}

\usepackage{amssymb}
\usepackage{amsmath,amsfonts}
\usepackage{amsopn,bm,mathtools}

\usepackage{nicefrac}       
\newcommand{\vct}[1]{\boldsymbol{#1}} 
\newcommand{\mat}[1]{\boldsymbol{#1}} 

\newcommand{\ProbOpr}[1]{\mathbb{#1}}

\newcommand{\expect}[2]{%
\ifthenelse{\equal{#2}{}}{\ProbOpr{E}_{#1}}
{\ifthenelse{\equal{#1}{}}{\ProbOpr{E}\left[#2\right]}{\ProbOpr{E}_{#1}\left[#2\right]}}} 
\newcommand{\var}[2]{%
\ifthenelse{\equal{#2}{}}{\ProbOpr{VAR}_{#1}}
{\ifthenelse{\equal{#1}{}}{\ProbOpr{VAR}\left[#2\right]}{\ProbOpr{VAR}_{#1}\left[#2\right]}}} 

\newcommand{\va}{\vct{a}}
\newcommand{\vb}{\vct{b}}
\newcommand{\vc}{\vct{c}}
\newcommand{\ve}{\vct{e}}

\newcommand{\vu}{\vct{u}}
\newcommand{\vv}{\vct{v}}

\newcommand{\vx}{{\vct{x}}}
\newcommand{\vy}{\vct{y}}

\newcommand{\mC}{\mat{C}}

\newcommand{\mE}{\mat{E}}

\newcommand{\mU}{\mat{U}}

\newcommand{\sD}{\mathcal{D}}

\newcommand{\vphi}{\vct{\phi}}

\newcommand{\vtheta}{\vct{\theta}}

\newcommand{\ourmethod}{{\textsc{Composer}}\xspace}
\newcommand{\ourtitle}{{Visually Grounded Concept Composition}}

\usepackage[colorinlistoftodos,linecolor=blue,backgroundcolor=white,bordercolor=blue,textsize=tiny,textwidth=20mm]
{todonotes}

\usepackage{qtree}
\usepackage{tabularx}
\usepackage{adjustbox}
\usepackage{tikz}

\usetikzlibrary{shapes,snakes}
\usepackage{xcolor}
\definecolor{codegreen}{rgb}{0,0.6,0}
\definecolor{codegray}{rgb}{0.5,0.5,0.5}
\definecolor{codepurple}{rgb}{0.58,0,0.82}
\definecolor{backcolour}{rgb}{0.95,0.95,0.92}
\definecolor{darkgreen}{rgb}{0,0.4,0}
\definecolor{cerise}{rgb}{0.871, 0.192, 0.388}
\definecolor{carmine}{rgb}{0.59, 0.0, 0.09}
\definecolor{olive}{rgb}{0.332, 0.418, 0.184}
\definecolor{navyblue}{rgb}{0.496, 0.810, 0.837}

\newcommand{\VMarker}{\raisebox{0.5pt}{\tikz\fill[codegreen] (0,0) circle (.5ex);}}

\newcommand{\VilMarker}{\raisebox{0.5pt}{\tikz\draw[codepurple,fill=codepurple] (0,0)--(1.154ex,0)--(.577ex,1ex)--(0,0);}}
\newcommand{\CMarker}{\raisebox{0.5pt}{\tikz\draw[cerise,fill=cerise] (0,.6ex)--(.6ex,0)--(1.2ex,.6ex)--(.6ex,1.2ex)--(0,.6ex);}}

\usepackage{amssymb}
\usepackage{pifont}
\newcommand{\cmark}{\text{\ding{51}}}
\newcommand{\xmark}{\text{\ding{55}}}

\newcommand{\suppl}{Appendix}

\usepackage{xspace}

\makeatletter
\DeclareRobustCommand\onedot{\futurelet\@let@token\@onedot}
\def\@onedot{\ifx\@let@token.\else.\null\fi\xspace}
\def\eg{\emph{e.g}\onedot} 
\def\ie{\emph{i.e}\onedot}

\def\wrt{w.r.t\onedot} 

\makeatother

\usepackage{mathtools}

\usepackage[ruled,vlined]{algorithm2e}

\usepackage{listings}

\graphicspath{{../figs/}{../}}

\newcommand{\eat}[1]{}

\usepackage{makecell}

\begin{document}
\title{\ourtitle}

\author{
  Bowen Zhang \footremember{Google}{Part of work done while at Google} \\
  USC \\\And
  Hexiang Hu   \footremember{USC}{Part of work done while at USC}\\
  Google Research  \\\And
  Linlu Qiu  \footremember{GoogleAIRes}{Work done as a Google AI resident.} \\
  Google Research  \\\AND
  Peter Shaw \\
  Google Research  \\\And
  Fei Sha \\
  Google Research 
\\}

\date{}

\maketitle

\begin{abstract}
We investigate ways to compose complex concepts in texts from primitive ones while grounding them in images. We propose Concept and Relation Graph (CRG), which builds on top of constituency analysis and consists of recursively combined concepts with predicate functions. Meanwhile, we propose a concept composition neural network called Composer to leverage the CRG for  visually grounded concept learning. Specifically, we learn the grounding of both primitive and all composed concepts by aligning them to images and show that learning to compose leads to more robust grounding results, measured in text-to-image matching accuracy. Notably, our model can model grounded concepts forming at both the finer-grained sentence level and the coarser-grained intermediate level (or word-level). Composer leads to pronounced improvement in  matching accuracy when the evaluation data has significant compound divergence from the training data.
\end{abstract}

\section{Introduction}
\label{sec:intro}

Visually grounded text expressions denote the images they describe. These expressions of visual concepts are naturally organized hierarchically in sub-expressions. The organization reveals structural relations that do not manifest when the sub-expressions are studied in isolation. For example, the phrase ``a soccer ball in a gift-box'' is a compound of two shorter phrases, \ie, ``a soccer ball'' and ``a gift-box'', but carries the meaning of the spatial relationship ``something in something'' that goes beyond the two shorter phrases separately. The compositional structure of the grounded expression requires a concept learner to understand what primitive concepts are visually appearing and how the compound relating multiple primitives modifies their appearance.

Existing approaches~\citep{kiros2014UVS,faghri2017vse++,Lu2019ViLBERTPT,chen2020uniter,chen2021vseinfty} tackle visual grounding via end-to-end learning, which typically learns to align image and text information using neural networks without explicitly modeling their compositional structures. While neural networks have shown strong generalization capabilities in test examples that are i.i.d to the training distribution~\cite{Devlin2019BERT}, they often struggle in dealing with out-of-domain examples of novel compositional structures, in many tasks such as Visual Reasoning~\citep{johnson2017clevr,bahdanau2018systematic,pezzelle2019MALeViC}, Semantic Parsing~\citep{finegan2018improving,keysers2020measuring}, and (Grounded) Command Following~\citep{lake2018generalization,chaplot2018gated,hermann2017grounded,ruis2020gscan}.

\begin{figure}[t]
    \centering
    \includegraphics[width=0.48\textwidth]{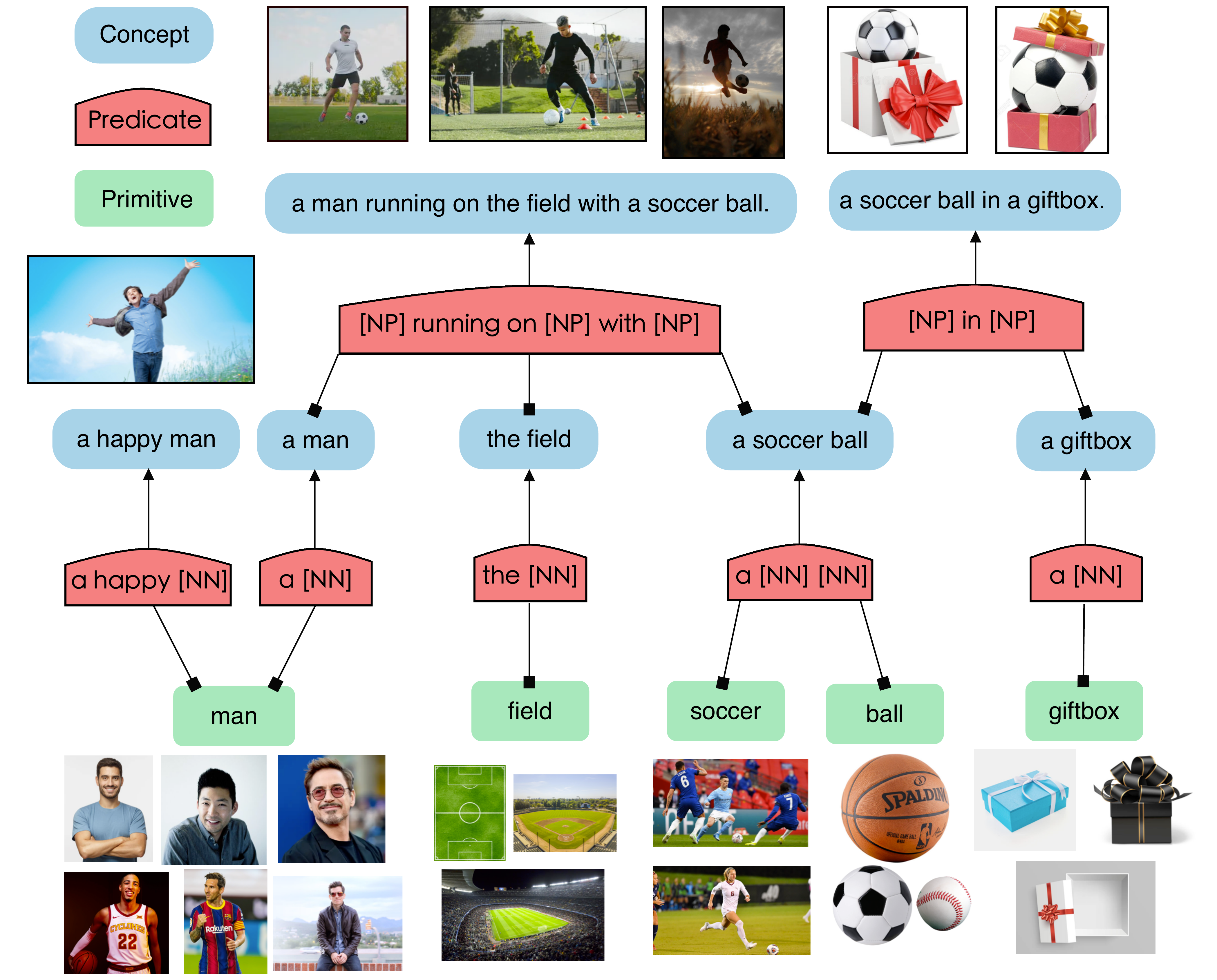}
    \caption{Concepts and their visual denotations organized by the Concept \& Relation Graph}
    \label{fig:crg}
\end{figure}

In this work, we investigate how complex concepts, composed of simpler ones, are grounded in images at sentences, phrases and tokens levels. In particular, we investigate whether the structures of how these concepts are composed can be exploited as a modeling prior to improve visual grounding. To this end, we design Concept \& Relation Graph (CRG), which is derived from constituency parse trees.
The resulting CRG is a graph-structured database where concept nodes encode language expressions of concepts and their visual denotations (\eg, a set of images corresponding to the concept), and predicate nodes define how a concept is semantically composed from its child concepts. Our graph is related to the denotation graph~\cite{young2014image,zhang2020learning} but differs in two key aspects. First, our graph extracts the concepts without specially crafted heuristic rules\footnote{Our graph construction relies on constituency parsing thus it is more scalable than hand-written rules initially developed for denotation graphs. The technique of denotation graph has been developed and evaluated on English language corpus, and its multilingual utility depends on the parsing techniques for those languages other than English.}. Secondly, CRG's predicate can encode richer information explicitly than the subsumption relationships implicitly expressed in the denotation graphs.
An illustrative figure of the graph is shown in Figure~\ref{fig:crg}.

In addition to CRG, we propose \textbf{C}oncept c\textsc{\textbf{ompos}}ition transform\textsc{\textbf{er}} (\ourmethod) that leverages the structure of text expressions to recursively encode the grounded concept embeddings, from coarse-level such as the noun words that refer to objects, to finer-grained ones with multiple levels of compositions. Transformer~\cite{Vaswani2017Transformer} is used as a building block in our model, to encode the predicates, and perform grounded concept composition. We learn \ourmethod using the task of visual-semantic alignment. Unlike traditional approaches, we perform hierarchical learning of visual-semantic alignment, which aligns the image to words, phrases, and sentences, and preserves the order of matching confidences.

We conduct experiments on multi-modal matching and show that \ourmethod achieves strong grounding capability in both sentence-to-image and phrase-to-image retrieval on the popular benchmarks. We validate the generalization capability of \ourmethod by designing an evaluation procedure for a more challenging compositional generalization task that uses test examples with maximum compound divergence (MCD) to the training data~\citep{shaw2020compositional,keysers2020measuring}. Experiments show that \ourmethod is more robust to the compositional generalization than other approaches.

\vspace{2pt}
\noindent\textbf{Our contributions} are summarized as below:
\begin{itemize}[leftmargin=*,topsep=1pt,itemsep=0pt]
    \item We study the compositional structure of visually grounded concepts and design Concept \& Relation Graph that reflects such structures.
    \item We propose \textbf{C}oncept c\textsc{\textbf{ompos}}ition transform\textsc{\textbf{er}} (\ourmethod) that recursively composes concepts using the child concepts and the semantically meaningful rules, which leads to strong compositional generalization performances.
    \item We propose a new evaluation task to assess the model's compositional generalization performances on the task of text-to-image matching and conduct comprehensive experiments to evaluate both baseline models and \ourmethod.
\end{itemize}

\begin{figure*}[t]
    \centering
    \includegraphics[width=0.975\textwidth]{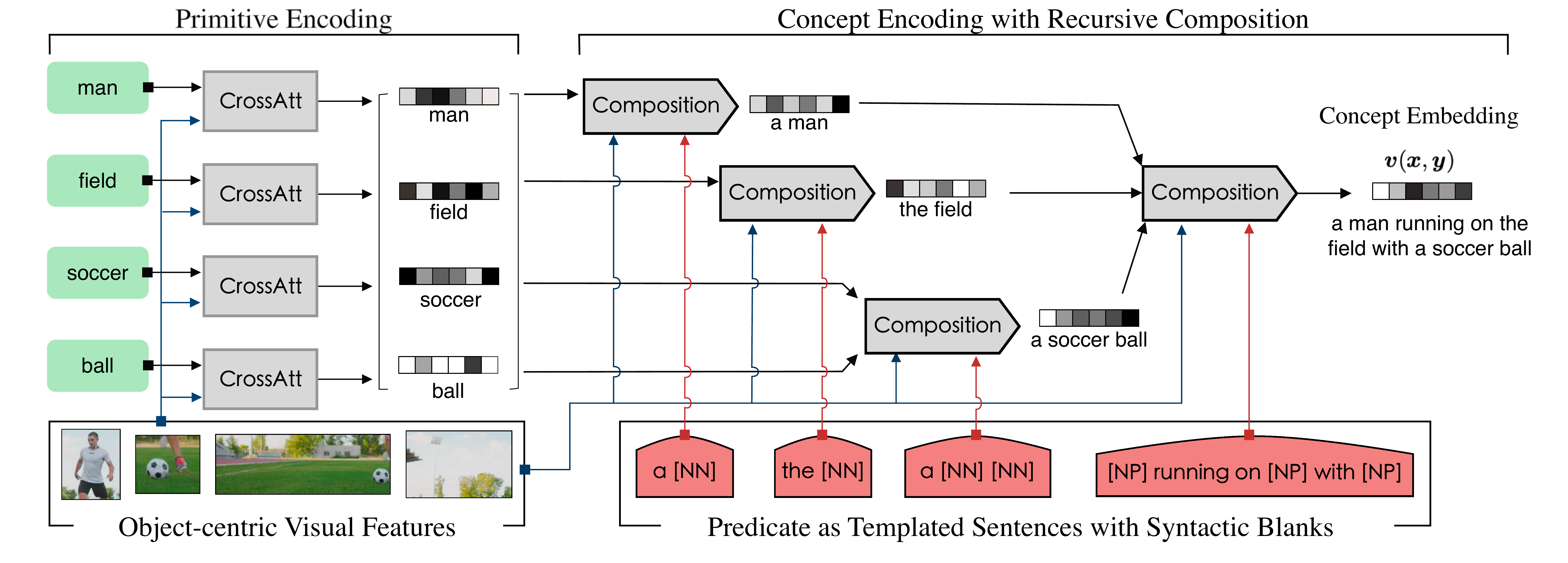}
    \caption{\small The overall design of the proposed \ourmethod model.}
    \label{fig:composer}
\end{figure*}

\section{Concept \& Relation Graph}
\label{sec:crg}

We introduce multi-modal Concept and Relation Graph (CRG), a graph composed of concept and predicate nodes, which compose visually grounded descriptive phrases and sentences.  Figure~\ref{fig:crg} provides an illustrative example. The concepts include sentences and intermediate phrases, shown as \textcolor{CornflowerBlue}{\bf blue nodes}. The primitives are the leaf nodes (typically noun words) that refer to visual objects, shown as \textcolor{LimeGreen}{\bf green nodes}.
The predicates (\textcolor{RedOrange}{\bf red nodes}) are $n$-ary functions that define the meaning of the concept composition. Their ``signatures'' consist of lexicalized templates, the number of arguments, and the syntactic type of the arguments. They combine primitives or simpler concepts into more complex ones.

\paragraph{Identifying concepts and relations.} Given pairs of aligned image and sentence, we first parse a sentence into a constituency tree, using a state-of-the-art syntactic parser~\cite{kitaev2018constituency}. We use the sentence's constituent tags to identify concepts and their relations. The set of relations are regarded as $n$-ary functions with placeholders denoted with constituency tags. We refer to such functions as predicates. Simpler concepts are arguments to the predicates, and the return values of the functions are complex concepts. The edges of the graph represent the relationship between predicates and their arguments. We restrict the type of constituents that can be concepts and how the predicates can be formed.

A concrete example is as follows: given an input concept ``two dogs running on the grass'', the algorithm extracts the predicate ``\texttt{[NP] running on [NP]}'' and the child concepts ``two dogs'' and ``the grass''. Here we use syntactic placeholders to replace the concept phrases. Details are in the \suppl. This idea is closely related to the semantically augmented parse trees~\citep{ge2009learning}, though we focus on visually grounded concepts.

\paragraph{Finding visually grounded concepts.} We take paired images and texts\footnote{In this paper, texts refer to sentences.}, and convert the texts into derived trees of predicates and primitives. With the generated text graph, we then group all images that refers to the same concept to form the image denotation, similar as \citet{young2014image} and \citet{zhang2020learning}. The image denotation is the set of images that contain the referred concept. For example, the image denotation of the concept “ball” is  all the images that have the visual object category “ball”. As a result, we associate the image denotation with each concept in the format of words, phrases, and sentences, which creates a multi-modal graph database as Figure~\ref{fig:crg}.

\section{\ourmethod: Recursive Modeling of the Compositional Structure}
\label{sec:method}

The main idea of \ourmethod is to recursively compose primitive concepts into sentences of complex structure, using composition rules defined by the predicates. Figure~\ref{fig:composer} presents a conceptual diagram of the high-level idea. Concretely, it first takes the primitive word embedding as the inputs and performs cross-modal attention to obtain their visually grounded word embeddings. Next, the \ourmethod calls the composition procedure to modify or combine primitive or intermediate concepts, according to the description of its predicates. At the end of this recursive procedure, we  obtain the desired sentence concept embedding. In the rest of this section, we first discuss the notation and backgrounds, then introduce how primitives and predicates are encoded (\S~\ref{subsec:encoding}), and present the recursive composition procedures in detail (\S~\ref{subsec:composition}). Finally, we discuss the learning objectives (\S~\ref{subsec:learning}).

\paragraph{Notation.} We denote a paired image and sentence as $(\vx, \vy)$ and the corresponding concepts and predicate for a tree $(\vx, \mU, \mE)$, where $\mU,\mE$ corresponds to the set of primitives and the set of predicates, respectfully. We also denote all concepts from a sentence $\vy$ to be $\mC$, where $\mU \not\subset \mC$ and $\vy \in \mC$.

\paragraph{Multi-head attention mechanism.}
Multi-Head Attention (MHA)~\citep{Vaswani2017Transformer} is the building block of our model. It takes three sets of input elements, \ie, the key set $K$, the query set $Q$, and the value set $V$, and perform scaled dot-product attention as:
\begin{equation}
    \texttt{MHA}(K, Q, V) = \texttt{FFN}\big(\texttt{Softmax}(\frac{Q^\top K}{\sqrt{d}}) \cdot V \big) \nonumber
\end{equation}
Here, $d$ is the dimension of elements in $K$ and $Q$. \texttt{FFN} is a feed-forward neural network. With different choices of $K$ and $V$, MHA can be categorized as self-attention (\texttt{SelfAtt}) and cross-attention (\texttt{CrossAtt}), which corresponds to the variants with $K$ and $V$ including only the single-modality or cross-modality features.

\subsection{Encoding Primitives and Predicates}
\label{subsec:encoding}

Given a paired image and sentence $(\vx, \vy)$, we parse the sentence as the tree of primitives and predicates $(\vx, \mU, \mE)$. Here, we represent the image as a set of visual feature vectors $\{\vphi\}$, which are the object-centric features from an object detector~\citep{anderson2017updown}. Noted that we didn’t use structural information beyond object proposals/regions. Our {\ourmethod} takes the primitives and predicates as input and output the visually grounded concept embeddings, with both the primitives and predicates as continuous vectors of different contextualization.

\paragraph{Representing primitives with visual context.}
The primitive concepts refer to tokens which can be visually grounded, and we represent them as word embeddings contextualized with visual features. As such, we use a one-layer Transformer with the $\texttt{CrossAtt}$ mechanism, where $K$, $V$, and $Q$ are linear transformations of $\vphi$, $\vphi$, and $\vu$, respectively. This essentially uses the word embedding to query the visual features and outputs the grounded primitive embeddings $\hat{\mU} = \{ \hat{\vu} \}$. Note that the output is always a single vector for each primitive as it is a single word.

\paragraph{Representing predicates as neural templates.} A predicate $\ve$ is a semantic $n$-place function that combines multiple concepts into one. We represent it as a \textbf{template sentence} with words and syntactic placeholders, such as ``\texttt{[NP]$_1$} running on \texttt{[NP]$_2$}'', where those syntactic placeholders denote the positions and types of arguments. We encode such template sentences via \texttt{SelfAtt} mechanism, using a multi-layer Predicate Transformer (PT). The output of this model is a contextualized sequence of the words and syntactic placeholders as $\hat{\ve}$.

\begin{figure}[t]
    \centering
    \includegraphics[width=0.5\textwidth]{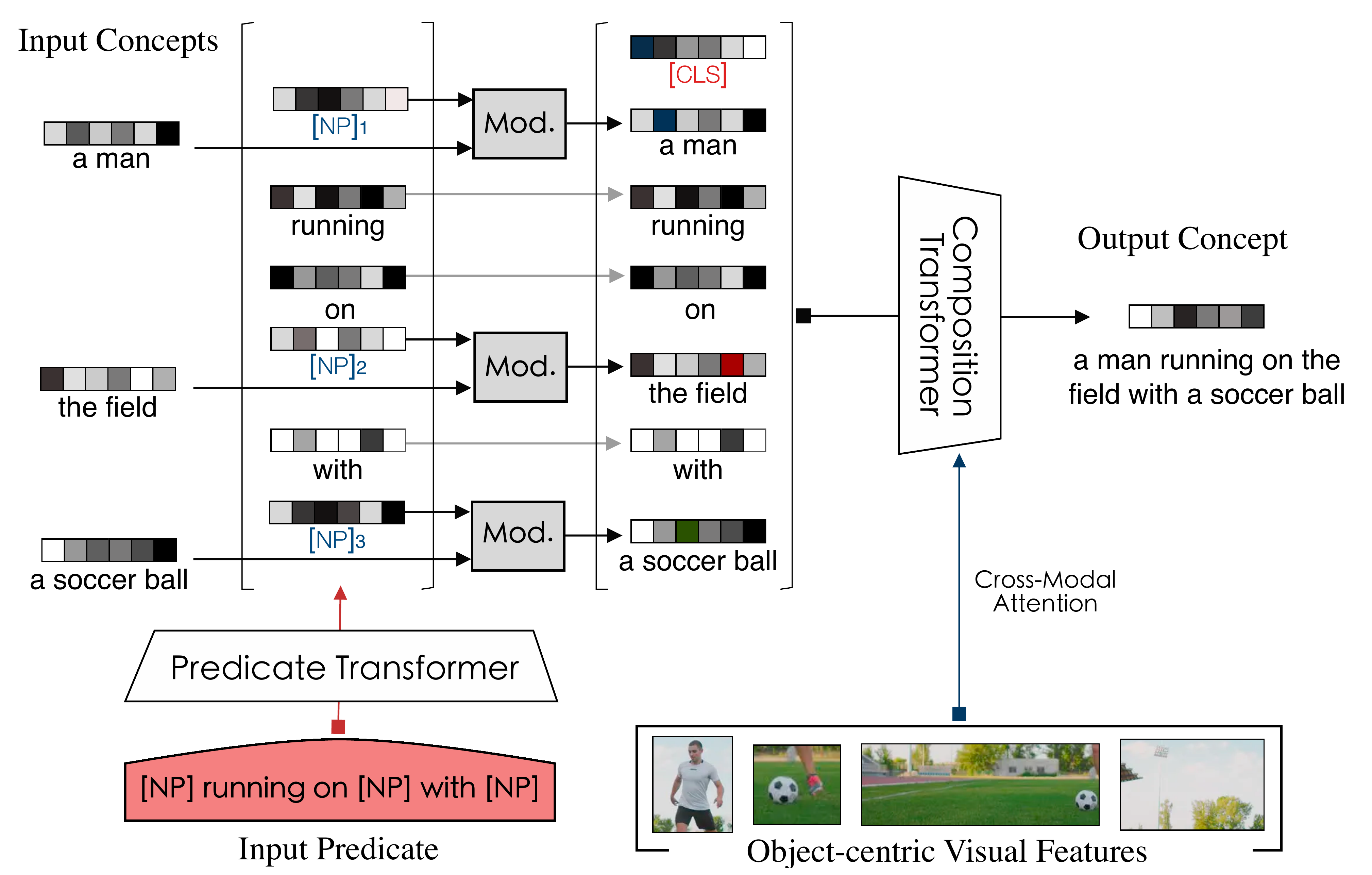}
    \caption{Details of the composition procedure.}
    \label{fig:composition}
    \vspace{-1em}
\end{figure}

\subsection{Recursive Concept Composition}
\label{subsec:composition}

With the encoded primitives $\hat{\mU}$ and predicates $\hat{\mE}$, the {\ourmethod} then performs multiple recursive composition steps to obtain the grounded concept embedding, $\vv{(\vx, \vy)}$, representing the visual-linguistic embedding of the sentence and the image as shown in the Figure~\ref{fig:composer}. To further illustrate this process, we detail the composition function in below, as shown in Figure~\ref{fig:composition}.

\paragraph{Input concept modulation.} 
We use a modulator to bind the arguments in the predicate to the input child concepts. Given a encoded predicate $\hat{\ve}=$ $\{\texttt{[NP]}_1$, \texttt{running}, \texttt{on}, $\texttt{[NP]}_2$, \texttt{with}, $\texttt{[NP]}_3\}$ and a input concept $\vc_1=$ \texttt{``a man''}, the modulator is a neural network that takes the concept embedding $\vc_1$ and its corresponding syntactic placeholder $\texttt{[NP]}_1$ as input and outputs a modulated embedding. This embedding is then reassembled with the embeddings of non-arguments in the predicate and used for the later stage.
For example, the output sequence becomes $\{\texttt{Mod}(\texttt{[NP]}_1, \vc_1)$, \texttt{running}, \texttt{on}, $\texttt{Mod}(\texttt{[NP]}_2, \vc_2)$, \texttt{with}, $\texttt{Mod}(\texttt{[NP]}_3, \vc_3)\}$ after the modulator processed each pair of input concept and syntactic placeholder.
Various choices of neural networks are available for this modulator, such as a Multi-Layer Perceptron (MLP) or a Feature-wise Linear Modulation (FiLM)~\cite{perez2018film}. \ourmethod uses FiLM for its strong empirical performance.

\paragraph{Contextualization with visual context.} After concept modulation, we get a sequence of embeddings for non-argument words of the predicate and the binded child concepts, which is then fed as an input to a Composition Transformer (CT) model. This Transformer has multiple layers, with both \texttt{CrossAtt} layers that attends to the object-centric visual features and \texttt{SelfAtt} layers that contextualize between tokens. Please refer to {\suppl} for the detailed network architecture. 

Given that our model is recursive by nature, the computation complexity of CT is proportional to the depth of the tree. We provide a comprehensive study in \S~\ref{subsec:ablation} to show the correlation between the parameter/complexity and model's performances.

\subsection{Learning \ourmethod with Visual-Semantic Alignments}
\label{subsec:learning}

With the composed grounded concept embedding $\vv(\vx, \vy)$, we use the visual-semantic alignment as the primary objective to learn {\ourmethod}. To this end, we compute the alignment score by learning an additional linear regressor $\theta$:
\begin{equation}
    \vct{s}({\vx, \vy}) = \theta^\top \cdot \vv(\vx, \vy) \propto p(\vx, \vy), \nonumber
\end{equation}
where $p(\vx, \vy)$ is the probability that the sentence and image is a good match pair. Then we learn the sentence to image alignment by minimizing the negative log-likelihood (NLL):
\begin{align}
    \ell_{\textsc{match}} = - \sum_i \log \frac{\exp({s(\vx_i, \vy_i)})}{\sum_{(\hat{\vx}, \hat{\vy}) \sim \sD_i}\exp({s(\hat{\vx}, \hat{\vy})})} \nonumber
\end{align}
with $\sD_i = \{(\vx_i, \vy_i)\}\cup\sD^-_i$. To properly normalize the probability, it is necessary to sample a set of negative examples to contrast. Thus, we generate $\sD^-_i$ using the strategy of \citet{Lu2019ViLBERTPT}.

\paragraph{Multi-level visual-semantic alignment (MVSA).} Since \ourmethod composes grounded concepts recursively from the primitives, we obtain the embeddings of all the intermediate concepts automatically. Therefore, it is natural to extend the alignment learning objectives to all those intermediate concepts. We optimize the triplet hinge loss~\citep{kiros2014UVS}:
\begin{align}
    \ell_{\textsc{MVSA}} = & \sum_i \sum_{\vc \in \mC_i} [\alpha - {s}{(\vx_i, \vc)} + {s}{(\vx_i, {\vc}^-)}]_+ \nonumber \\
    & + [\alpha - {s}{(\vx_i, \vc)} + {s}{({\vx}^-_i, \vc)}]_+ \nonumber
\end{align}
where $[h]_+ = \text{max}(0, h)$ denotes the hinge loss and $\alpha$ is the margin to be tuned. We derive the negative concepts $\vc^-$ from the negative sentences in the $\sD^-_i$. We observe that negative concepts at word/phrase levels are noisier than the ones at sentence level because many are common objects presented in the positive image and lead to ambiguity in learning. Therefore, we choose hinge loss over NLL because it is more robust to label noises~\cite{biggio2011support}.

\paragraph{Learning to preserve orders in the tree.} Finally, we use an order-preserving objective proposed by \citet{zhang2020learning}, to ensure that a fine-grained concept (closer to sentence) can produce a more confident alignment score than a coarse-grained concept (closer to primitive):
\begin{equation}
    \ell_{\textsc{order}} = \sum_i \sum_{\ve_{jk}} [\beta - s(\vx_{i}, \vc_j) + s(\vx_{i}, \vc_k) ]_+ \nonumber
\end{equation}
Here, $\ve_{jk}$ represents a predicate connecting the $\vc_j$ and $\vc_k$, with $\vc_j$ to be the fine-grained parent concept which is closer to the sentence and $\vc_k$ to be the coarse-grained child concept which is closer to the primitives. $\beta$ is the margin that sets the constraint on how hard the order of embeddings should be reserved.

The complete learning objective is a weighted combination of three individual losses defined above, with the loss weights $\lambda_1 = 1$ and $\lambda_2 = 1$:
\begin{equation}
    \ell = \ell_{\textsc{match}} + \lambda_1 \cdot \ell_{\textsc{mvsa}} + \lambda_2 \cdot \ell_{\textsc{order}} \nonumber
\end{equation}
The details of model optimization and hyper-parameter setting are included in the \suppl.

\section{Related Work}
\label{sec:related}

\paragraph{Generalization in grounded language understanding.}
Many evaluation methods are proposed to assess the model's generalization capabilities in grounded language understanding. \citet{johnson2017clevr} proposes a synthetic dataset, \ie CLEVR, to evaluate the generalization of visual question answering models to novel objects and attributes. \citet{misra2017red} proposes to evaluate compositional generalization capability of visual models \wrt short phrases consist of attributes and objects.
\citet{chaplot2018gated} and \citet{hermann2017grounded} evaluate RL agents' capability to generalize to a novel composition of shape, size, and color in 3D simulators, which shows that RL agents generalize poorly. gSCAN~\cite{ruis2020gscan} perform a systematic benchmark to assess command following in a grounded environment. In this work, we focus on assessing model composition generalization under the visual context.

\paragraph{Compositional networks.}
State-of-the-art visually grounded language learning typically use deep Transformer models~\citep{Vaswani2017Transformer} such as ViLBERT \cite{Lu2019ViLBERTPT}, LXMERT \cite{tan2019lxmert} and UNITER \cite{chen2020uniter}. Though being effective for data over i.i.d distribution, these models do not explicitly exploit the structure of the language and are thus prone to fail on compositional generalization. In contrast, another thread of works \cite{andreas2016neural,yi2018neural,mao2019neuro,shi2019visually,wang2018scene} parse the language into an executable program composed as a graph of atomic neural modules, where each module is designed to perform atomic tasks and are learned end-to-end. Such models show almost perfect performances on synthetic benchmarks~\cite{johnson2017clevr} but perform subpar on the real-world data~\cite{young2014image,chen2015coco-cap} that are noisy and highly variable. Unlike them, we propose using a compositional neural network based on the Transformer architecture, which extends state-of-the-art neural networks to explicitly exploits language structure.

\section{Experiment}
\label{sec:exp}
In this section, we perform experiments to validate the proposed \ourmethod model on the tasks of sentence-to-image retrieval and phrase-to-image retrieval. We begin with introducing the setup in \S~\ref{subsec:setup} and then present the main results in \S~\ref{subsec:main}, comparing models for their in-domain, cross-dataset evaluation, and compositional generalization performance. Finally, we perform an analysis and ablation study of our model design in \S~\ref{subsec:ablation}.

\begin{table}[tb]
    \centering
    \small
    {
        \tabcolsep 2pt
	    \begin{tabular}{@{\;} ccccc@{\;}}
	        \addlinespace
	        \toprule
	        Dataset & {\# concepts} & {\# predicates} & {\# primitives} & {Avg height} \\ 
	        \midrule
	        F30K & 408,464 & 122,196 & 10,755 & 3.09\\
    	    C30K & 345,331 & 88,623 & 9,683 & 2.86 \\
	        \bottomrule
	    \end{tabular}
	}
	\caption{Statistics of the concepts and predicates in the F30K and C30K datasets.}
    \label{tab:crg_stats}
\end{table}

\subsection{Experiment Setup}
\label{subsec:setup}

\paragraph{Datasets.} We perform experiments on the COCO-caption (COCO)~\cite{chen2015coco-cap} and Flickr30K (F30K)~\cite{young2014image} datasets. Each image of these two datasets is associated with five sentences. Flickr30K contains 31,000 images, and we use the same data split as~\cite{faghri2017vse++}, where there are 29,000 training images, 1000 test images, and 1000 validation images. COCO contains 123,287 images in total. For fast iteration, we use a subset training data C30K, which contains the same amount of images as the F30K. Note that C30K is a training split. We also trained models on the full COCO training split. For COCO dataset, the results are evaluated on COCO 1K test split~\cite{karpathy2015deep}.  We use COCO 1k test split for both in-domain (models trained on either C30K or full COCO training split and evaluate on COCO-caption) and cross-dataset transfer (models trained on F30K and evaluate on COCO-caption) evaluation. For both F30K and COCO 1K test split, there are 5,000 text queries and 1,000 candidate images to be retrieved. We report recall@1 (R1) and recall@5 (R5) as the primary retrieval metric.

\paragraph{Compositional generalization evaluation.} To generate evaluations of compositional generalization, we use a method similar to that of \citet{shaw2020compositional} and \citet{keysers2020measuring} which maximizes \emph{compound divergence} between the distribution of \emph{compounds} in the evaluation set and in the training set. Here compounds are defined based on the predicates occurring in captions. Following this method, we first calculate the overall divergence of compounds from the evaluation data to the training data using predicates from all the sentences. Then, for each sentence in the evaluation data, we calculate a compound divergence with this specific example removed. We rank those sentences based on the difference of the compound divergence. Finally, we choose the top-K sentences with the largest compound divergence differences and its corresponding images to form the evaluation splits.

Using this method, we generate evaluation splits with 1,000 images and 5,000 text queries, COCO-MCD and F30K-MCD, to assess models trained on F30K and COCO, respectively. Therefore, these splits assess both compositional generalization and cross-dataset transfer. Defining such splits across datasets is also helpful to achieve greater compound divergence than is otherwise possible, given the small amount of available in-domain test data. More details are included in \suppl.

\paragraph{CRG construction.} We constructed two CRGs on the F30K and C30K datasets, using the procedure mentioned in \S~\ref{sec:crg}. The key statistics of the graph we generated as shown in Table~\ref{tab:crg_stats}.

\paragraph{Baselines and our approach.} We compare \ourmethod to two strong baseline methods, \ie,  ViLBERT~\cite{Lu2019ViLBERTPT} and VSE~\cite{kiros2014UVS}. We make sure all models are using the same object-centric visual features extracted from the Up-Down object detector~\cite{anderson2017updown} for fair comparison. For the texts, both ViLBERT and the re-implemented VSE use the pre-trained BERT model as initialization. For the \ourmethod, we only initialize the predicate Transformer with the pre-trained BERT, which uses the first six layers. Note that the ViLBERT results are re-produced using the codebase from its author. ViLBERT is \textbf{not pre-trained} on any additional data of image-text pairs to prevent information leak in both cross-dataset evaluation and compositional generalization. Therefore, we used the pre-trained BERT models provided by HuggingFace to initalize the text stream of ViLBERT, and then followed the rest procedure in the original ViLBERT paper. Please refer to {\suppl} for complete details.

\begin{table}[t]
    \centering
    \small
    {
        \tabcolsep 2pt
	    \begin{tabular}{@{\;} l@{\;\;} cc@{\quad} cc@{\quad} cc@{\;}}
	        \addlinespace
	        \multicolumn{7}{c}{\textbf{(a)} Models trained on F30K} \\
	        \toprule
	        {Eval on} & \multicolumn{2}{c}{F30K} &
	        \multicolumn{2}{c}{COCO} & \multicolumn{2}{c}{COCO-MCD} \\
	        \cmidrule(lr){2-3} \cmidrule(lr){4-5} \cmidrule(lr){6-7}
	        {Method} & {R1} & {R5} & {R1} & {R5} & {R1} & {R5} \\
	        \midrule
             \VMarker{} VSE & 46.84 & 77.16 & 25.60 & 54.36 & 21.82 & 47.58 \\
             \VilMarker{} ViLBERT & 50.94 & \bf 80.86 & 30.50 & 58.98 & 24.44 & 51.44 \\
             \midrule
    	     \CMarker{} \ourmethod & \bf 54.02 & 80.27 & \bf 33.81 & \bf 63.19 & \bf 29.20 & \bf 57.13  \\
	        \bottomrule
            
            \addlinespace
            \multicolumn{7}{c}{\textbf{(b)} Models trained on C30K} \\
	        \toprule
	        {Eval on} & \multicolumn{2}{c}{COCO} &
	        \multicolumn{2}{c}{F30K} & \multicolumn{2}{c}{F30K-MCD} \\
	        \cmidrule(lr){2-3} \cmidrule(lr){4-5} \cmidrule(lr){6-7}
	        {Method}& {R1} & {R5} & {R1} & {R5} & {R1} & {R5} \\
	        \midrule
	        \VMarker{} VSE & 45.74 & \bf 81.22 & 27.66 & 55.92 & 23.44 & 47.90 \\
	        \VilMarker{} ViLBERT &\bf 48.08 & 81.10 & 31.12 & 58.88 & 24.02 & 49.34\\
	        \midrule
    	    \CMarker{} \ourmethod & 47.87 & 80.93 & \bf 34.29 & \bf 61.00 & \bf 26.91 & \bf 51.46 \\
	        \bottomrule
	    \end{tabular}
	}
	\caption{Text-to-Image retrieval results.}
    \label{tab:main}
\end{table}

\begin{figure}[t]
    \centering
    \begin{tabular}{@{}c@{}c@{}}
        \includegraphics[width=0.245\textwidth]{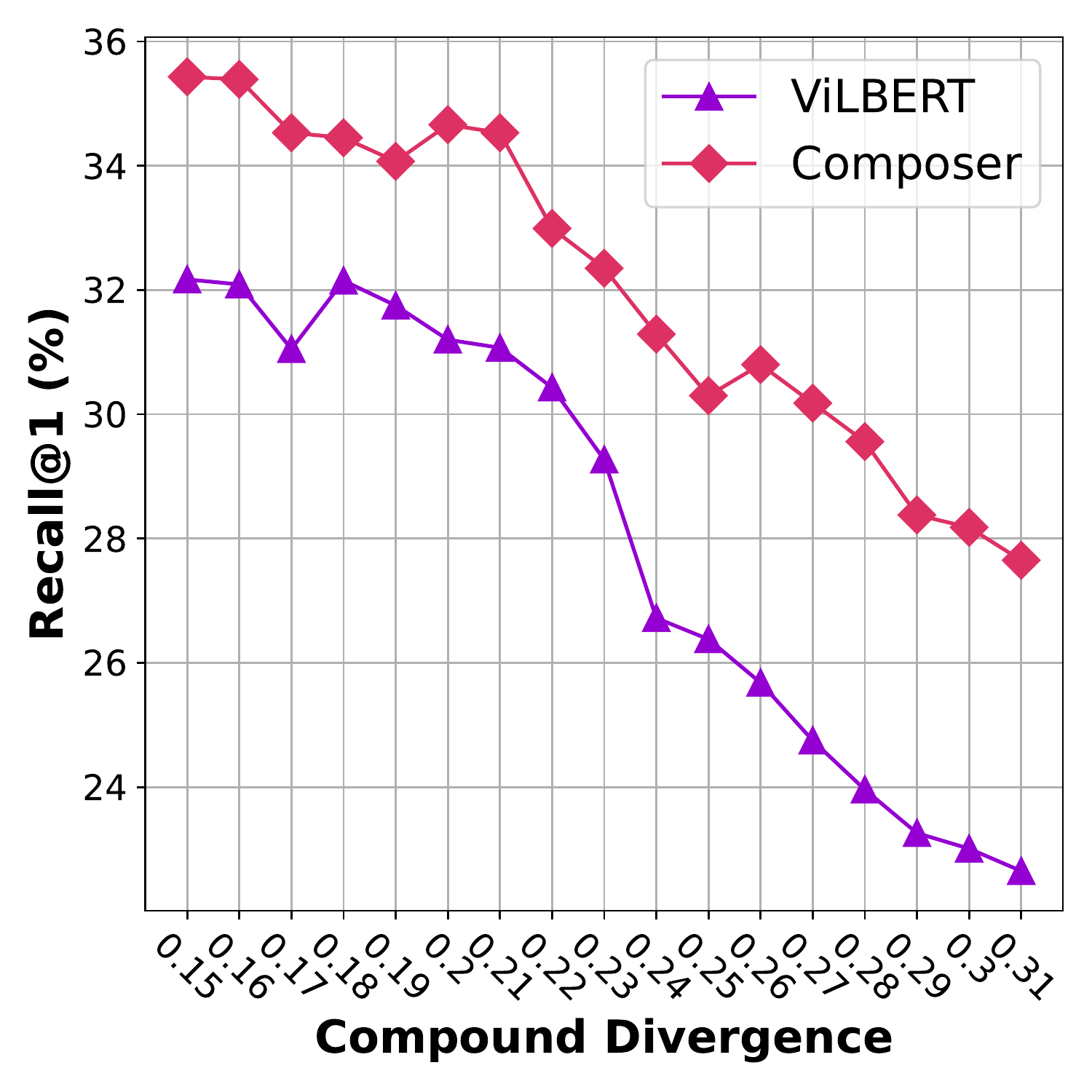} &
        \includegraphics[width=0.245\textwidth]{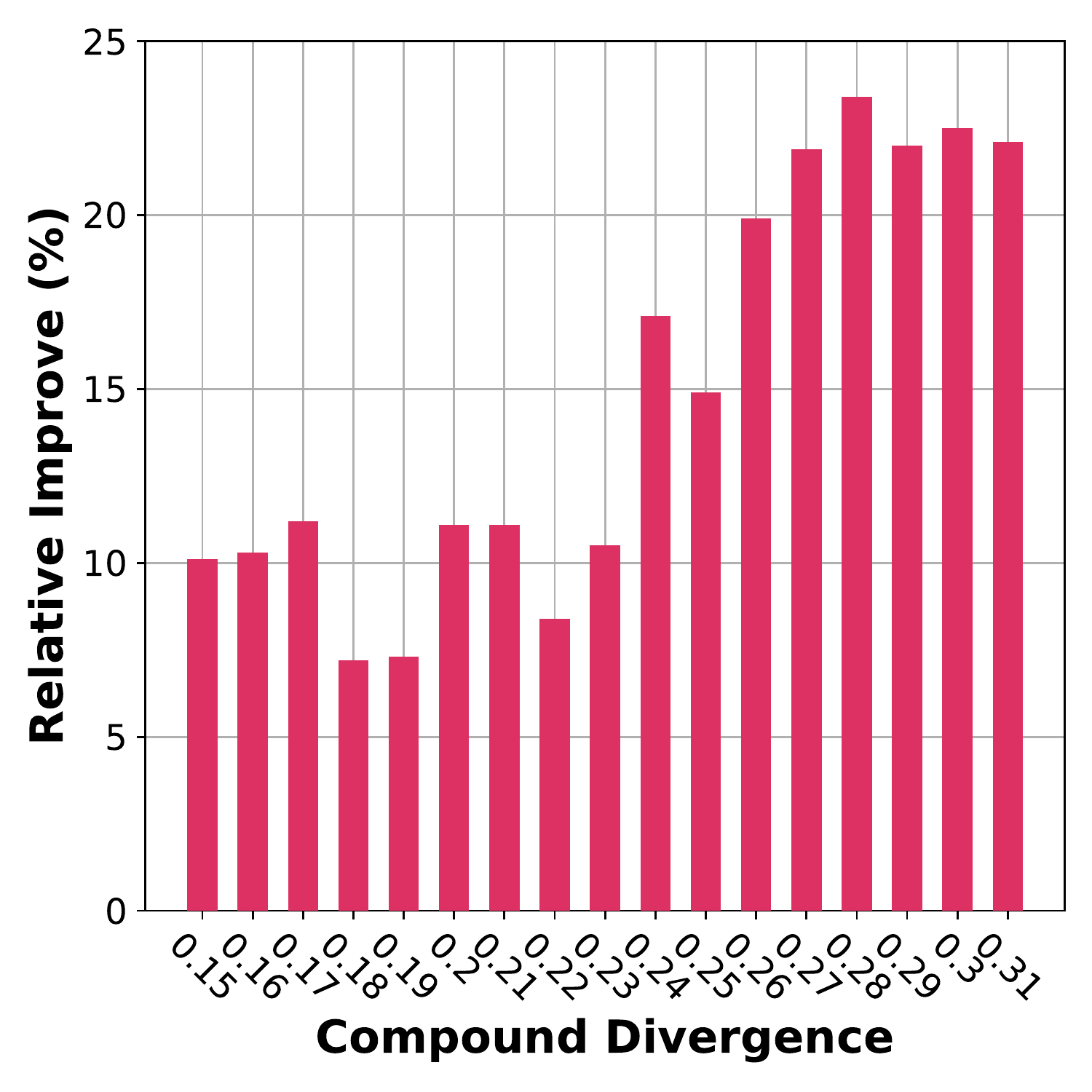}
    \end{tabular}
    \caption{\ourmethod's results on generalization splits of different compound divergence over text description (evaluated under the F30K$\rightarrow$COCO setting).}
    \vspace{-1em}
    \label{fig:compound_div}
\end{figure}

\subsection{Main Results}
\label{subsec:main}

We compare the {\ourmethod} with ViLBERT~\cite{Lu2019ViLBERTPT} and VSE~\cite{kiros2014UVS} on F30k and COCO for in-domain, zero-shot cross-dataset transfer, and compositional generalization (\eg F30K$\rightarrow$COCO-MCD). The notation A$\rightarrow$B means that the model is trained on A and evaluated on B. We report the results of sentence-to-image retrieval in the main paper and defer more ablation study results to the \suppl.

\paragraph{In-domain performance.}
Table~\ref{tab:main} presents the in-domain performance on both F30k and COCO datasets. First, we observe that both \ourmethod and ViLBERT consistently outperform VSE, which is expected as ViLBERT contains a cross-modal transformer with stronger modeling capacity. Comparing to ViLBERT, the {\ourmethod} performs on par.

\paragraph{Zero-shot cross-dataset transfer.}
We also consider zero-shot cross dataset transfer where we evaluate models on a dataset that is different from the training dataset. In this setting, the {\ourmethod} outperforms ViBLERT and VSE significantly. Concretely, on the F30k$\rightarrow$COCO setting, the {\ourmethod} improves R1 and R5 by $11.0\%$ and $7.0\%$ over the ViLBERT, relatively. There are $10.0\%$ and $4.2\%$ relative improvements on R1 and R5 on the other transfer direction.

\paragraph{Compositional generalization.}
On the max compound divergence (MCD) split, \ourmethod outperforms baselines by a margin for both F30K and C30K trained models (shown as Table~\ref{tab:main}). To further characterize the performance on compositional generalization, we create 16 test splits on each dataset with different compound divergence (from $0.15$ to $0.31$, where $0.31$ is the max CD) and present the results in Figure~\ref{fig:compound_div}. With the increases of CD, we observe the performance of \ourmethod and ViLBERT decreases. Compared to ViLBERT, we observe that \ourmethod is relatively more robust to this distribution shift, as the relative performance improvement is increasing with CD increases.

\subsection{Analysis and Ablation Study}
\label{subsec:ablation}

We perform several ablation studies to analyze {\ourmethod}, and provide qualitative results to demonstrate the model's interpretability.

\begin{table}[t]
    \centering
    \small
    {
        \tabcolsep 5pt
	    \begin{tabular}{@{\;} c@{\quad} cc@{\quad} cc@{\;}}
	        \addlinespace
	        \toprule
	         & \multicolumn{2}{c}{F30K$\rightarrow$F30K} &
	        \multicolumn{2}{c}{F30K$\rightarrow$COCO} \\
	        \cmidrule(lr){2-3} \cmidrule(lr){4-5}
	        \texttt{CrossAtt?} & {R1} & {R5} & {R1} & {R5} \\ 
	        \midrule
	        \xmark & 52.38 & 79.09 & 33.33 & 60.97 \\
    	    \cmark & \bf 54.02 & \bf 80.27 & \bf 33.81 & \bf 63.19 \\
	        \bottomrule
	    \end{tabular}
	}
	\caption{Study of different primitive encodings.}
    \label{tab:primitive}
\end{table}

\begin{table}[t]
    \centering
    \small
    {
        \tabcolsep 5pt
	    \begin{tabular}{@{\;} l@{\quad} cc@{\quad} cc@{\;}}
	        \addlinespace
	        \toprule
            & \multicolumn{2}{c}{F30K$\rightarrow$F30K} &
	        \multicolumn{2}{c}{F30K$\rightarrow$COCO} \\
	        \cmidrule(lr){2-3} \cmidrule(lr){4-5}
	        {Modulation} & {R1} & {R5} & {R1} & {R5} \\ 
	        \midrule
	        \texttt{Replace} & 52.84 & 79.79 & 32.63 & 61.61 \\
	        \texttt{MLP} & 52.92 & 79.89 & 33.39 & 61.41 \\
    	    \texttt{FiLM} & \bf 54.02 & \bf	80.27 & \bf 33.81 & \bf 63.19 \\
	        \bottomrule
	    \end{tabular}
	}
    \caption{Study of different modulators.}
    \label{tab:predicate_type}
\end{table}


\begin{table}[t]
    \centering
    \small
    {
        \tabcolsep 5pt
	    \begin{tabular}{@{\;} l@{\quad} cc@{\quad} cc@{\;}}
	        \addlinespace
	        \toprule
	         & \multicolumn{2}{c}{F30K$\rightarrow$F30K} &
	        \multicolumn{2}{c}{F30K$\rightarrow$COCO} \\
	        \cmidrule(lr){2-3} \cmidrule(lr){4-5}
	        {Method} & {Sentence} & {Phrase} & {Sentence} & Phrase \\ 
	        \midrule
	        {\VilMarker{} ViLBERT} & 50.94 & 18.34 & 30.50 & 15.00  \\
	        \bf\xspace\xspace$+$ MVSA & 48.90 & \bf 23.55 & 29.90 & 18.73 \\
	        \midrule
	        {\CMarker{} \ourmethod} & 52.52 & 21.04 & 32.87 & 18.29 \\ 
    	    \bf\xspace\xspace$+$ MVSA & \bf 54.02 & 22.70 & \bf 33.81 & \bf 18.81 \\
	        \bottomrule
	    \end{tabular}
	}
    \caption{Comparison between ViLBERT and \ourmethod on multi-level visual-semantic alignment supervision (MVSA). All results are reported in R1.}
    \label{tab:dense}
\end{table}

\begin{figure*}[t]
    \centering
    \includegraphics[width=0.985\textwidth]{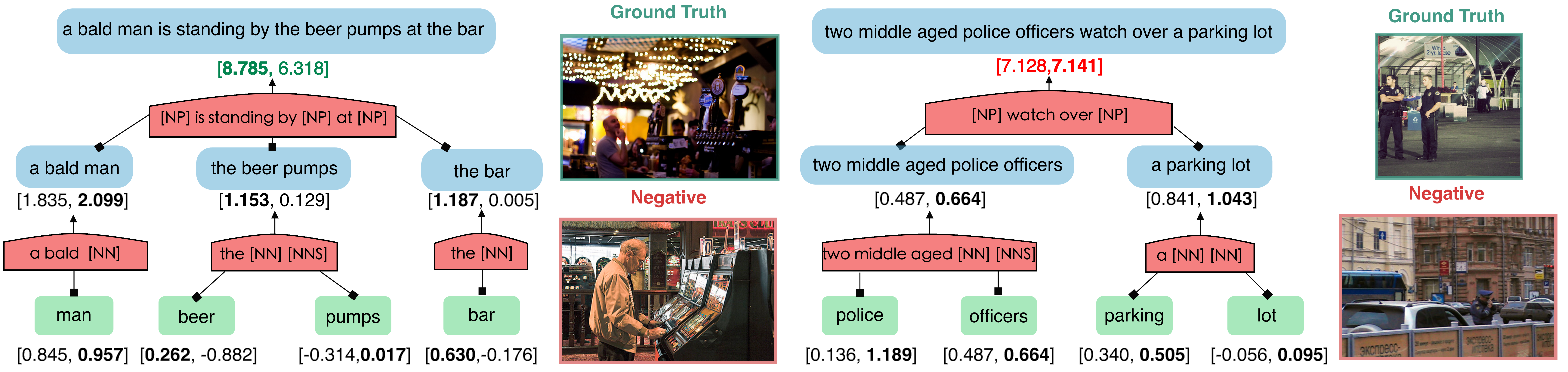}
    \caption{Interpreting the \ourmethod using visual-semantic alignment scores, formatted as $\texttt{[}\vct{s}_{\texttt{GT}}, \vct{s}_{\sc \texttt{Negative}}\texttt{]}$. The left figure corresponds to a correct example, and the right figure corresponds to an incorrect one.}
    \label{fig:interpretation}
    \vspace{-1em}
\end{figure*}

\paragraph{Is \texttt{CrossAtt} in primitive encoding useful?}
Table~\ref{tab:primitive} compares variants of {\ourmethod} with and without \texttt{CrossAtt} for primitive encoding, and shows that \texttt{CrossAtt} improves all metrics in in-domain and cross-dataset evaluation.

\paragraph{Which modulator works better?}
We consider three modulators to combine input concepts with the syntax token embeddings for later composition, which are \texttt{Replace}, \texttt{MLP}, and \texttt{FiLM}. The \texttt{Replace} directly replaces the syntax embedding with the input concept embedding. This is an inferior approach by design as it ignores the relative position of each concept. \texttt{MLP} model applies multi-layer neural networks on the concatenated syntax and input concept embeddings. \texttt{FiLM} model uses the syntax embedding to infer the parameter of an affine transformation, which is then applied to the input concepts. We show the results in Table~\ref{tab:predicate_type}. {\texttt{Replace}} achieves the worst performance, indicating the importance of identifying the position of input concepts. \ourmethod chooses {\texttt{FiLM}} as the modulator given its strong performance over all metrics.

\paragraph{Is MVSA supervision useful?}
We evaluate the influence of multi-level visual-semantic alignment on sentence and phrase to image retrieval. In the phrase-to-image experiments, we sample 5 non-sentence concepts from the CRG for each annotation in the corresponding test data and use them as the query to report results (in R1). Table~\ref{tab:dense} presents the results.
With the MVSA, \ourmethod outperforms ViLBERT on both sentence and phrase-based retrieval by a noticeable margin, indicating the advantage of capturing mid-level alignment in our model design.
Secondly, MVSA improves both \ourmethod and ViLBERT on the phrase to image retrieval over their counterparts. However, adding MVSA on ViLBERT leads to a degradation of sentence-to-image retrieval, showing that ViLBERT is incapable of mastering visual alignments for both sentences and phrases simultaneously. \ourmethod with MVSA improves itself on both sentence and phrase, showing strong multi-granular visual-semantic alignment ability.


\begin{table}[t]
    \centering
    \small
    {
        \tabcolsep 2.5pt
	    \begin{tabular}{@{\;} c@{\;}c@{\quad} cccc@{\;\;}cc@{\;}}
	        \addlinespace
	        \toprule
	        & & \multicolumn{2}{c}{F30K$\rightarrow$F30K} &
	        \multicolumn{2}{c}{F30K$\rightarrow$COCO} \\
	        \cmidrule(lr){3-4} \cmidrule(lr){5-6}
	        {PT} & {CT} & {R1} & {R5} & {R1} & {R5} & {\# Param} & {FLOPS} \\
	        \midrule
	        \texttt{2} & \texttt{5} & 51.72 & 79.71 & 33.67 & 60.38 & $129M$ & $35.40G$ \\
	        \texttt{4} & \texttt{5} & 53.32 & 79.73 & \bf 33.83 & 61.61 & $143M$ & $37.11G$ \\
    	    \texttt{6} & \texttt{5} & \bf 54.02 & 80.27 & 33.81 & \bf 63.19 & $157M$ & $38.40G$ \\
    	    \texttt{6} & \texttt{3} & 47.92 & 76.85 & 25.74 & 51.45 & $136M$ & $29.40G$ \\
    	    \texttt{6} & \texttt{1} & 34.47 & 62.61 & 21.25 & 43.86 & $115M$ & $19.98G$ \\
    	    \midrule 
    	    \multicolumn{2}{c}{ViLBERT} & 50.94 & \bf 80.86 & 30.50 & 58.98& $235M$ & $24.44G$ \\
	        \bottomrule
	    \end{tabular}
	}
    \caption{Results on \ourmethod of different complexity. All results are reported in R1. (PT: Predicate Transformer, CT: Composition Transformer)}
    \label{tab:complexity}
\end{table}

\paragraph{Performance vs. complexity trade-off.}
We compare variants of \ourmethod with different parameter and computation budgets, which uses different numbers of layers for the Predicate Transformer (PT) and Composition Transformer (CT). The results are shown in Table~\ref{tab:complexity}. First, We keep the size of CT fixed and vary the size of PT. It shows a marginal performance decrease occurring as the \# of layers of PT goes down. Then we keep the size of PT fixed and decrease the capacity of CT, which presents a significant performance drop, showing the essential role CT is playing. Besides having superior results, \ourmethod has \textit{(at least 33\%) fewer parameters} than the ViLBERT model, which indicates a potential performance gain could be achieved with a larger \ourmethod model.

For computation complexity, we observe that the full \ourmethod model is 50\% less efficient to a ViLBERT model, due to its recursive nature. Meanwhile, we notice that the increase in the \# of CT layers contributes a significant amount to the total computation time as every two additional layers adds $\sim10G$ FLOPS.

\paragraph{Performance under different parsing qualities.}
CRG is generated based on constituent parser. We investigate the performance of \ourmethod with CRG under different parsing qualites. Given a parsing tree, We randomly remove its branches randomly with a probability of 0.1, 0.3, or 0.5 to generate a tree with degraded parsing quality. We evaluate \ourmethod on the resulting CRGs. We summarized the results in Table~\ref{tab:parsing_qualities}. When parsing quality drops, both in-domain and cross-dataset transfer performance drops. The performance degrades by 12\%, when half of the parse could be missing. We expect with better parsing quality, \ourmethod can achieve stronger performance.

\begin{table}[t]
    \centering
    \small
    {
        \tabcolsep 5pt
	    \begin{tabular}{@{\;} c@{\quad} cc@{\quad} cc@{\;}}
	        \addlinespace
	        \toprule
	         & \multicolumn{2}{c}{F30K$\rightarrow$F30K} &
	        \multicolumn{2}{c}{F30K$\rightarrow$COCO} \\
	        \cmidrule(lr){2-3} \cmidrule(lr){4-5}
	        Pruning Probability & {R1} & {R5} & {R1} & {R5} \\ 
	        \midrule
	        Un-pruned &  \bf 54.02 & \bf 80.27 & \bf 33.81 & \bf 63.19  \\
    	    Probability=0.1 & 49.12 & 76.92 & 31.12 & 58.86 \\
    	    Probability=0.3 & 48.46 & 76.60 & 30.40 & 58.16 \\
    	    Probability=0.5 & 47.44 & 76.24 & 30.38 & 57.62 \\
	        \bottomrule
	    \end{tabular}
	}
	\caption{Performance under different parsing qualities.}
    \label{tab:parsing_qualities}
\end{table}

\paragraph{Interpreting \ourmethod's decision.}
Despite the solid performance, \ourmethod is also highly interpretable. Specifically, we visualize its alignment scores along with the concept composition procedure in Figure~\ref{fig:interpretation}. Empirically, we observe that most failures are caused by visually grounding mistakes at the primitive concepts level. The error then propagates “upwards” towards concept composition. 

For instance, the left example shows that \ourmethod is confusing between the ground truth and negative image when only the text of shared visual concept ``a bold man'' is presented. With more information are given, it gets clarified immediately as it notices that the target sentence is composed not only with the above subject, but also with the prepositional phrases ``by the beer pumps at the bar'' that reflects the visual environment.

\paragraph{Scalability to full COCO dataset.} Finally, we trained our model (PT=6, CT=5) on the full COCO training split and evaluated for both in-domain and cross-dataset transfer task. We use the same hyperparameters as C30K. However, \ourmethod underperforms the ViLBERT in this setting, as it achieves 56.06\% and 44.24\% in R1 for the in-domain task (COCO$\rightarrow$COCO) and cross-dataset evaluation tasks (COCO$\rightarrow$F30k), while ViLBERT obtains 56.83\% and 46.62\%, respectively.
We hypothesize that this negative result is largely due to the limited model capacity of the proposed \ourmethod, as it has relatively $33\%$ less parameters comparing to ViLBERT. Meanwhile, it is also observed that {\ourmethod} performs worse than ViLBERT in fitting training data. We observe that doubling the training epoch would increase both in-domain and out-of-domain performance by 2\% relatively. Increasing the layer of Composition Transformer (CT) to 7 would also improves R1 by 2.5\% relatively. Further scaling up \ourmethod may resolve this issue but requires more computational resources, and we leave this for future research.

\section{Conclusion}
\label{sec:conclusion}

In this paper, we propose the concept and relation graph (CRG) to explore the compositional structure in visually grounded text data. We further develop a novel concept composition neural network ({\ourmethod}) on top of the CRG, which leverages the explicit structure to compose concepts from word-level to sentence-level.
We conduct extensive experiments to validate our model on image-text matching benchmarks. Comparing with prior methods, {\ourmethod} achieves significant improvements, particularly in zero-shot cross-dataset transfer and compositional generalization. Despite these highlights, there are also many challenges that {\ourmethod} does not address in the scope of this paper. First, it requires high-quality parsing results to achieve strong performances, which may not be readily available in languages beyond English. Moreover, similar to other recursive neural networks, {\ourmethod} is also computationally resource demanding, which sets a limit to its scalability to large-scale data.

\section*{Acknowledgements} This work is partially supported by NSF Awards IIS-1513966/ 1632803/1833137, CCF-1139148, DARPA Award\#: FA8750-18-2-0117,  FA8750-19-1-0504, DARPA-D3M - Award UCB-00009528, Google Research Awards, gifts from Facebook and Netflix, and ARO\# W911NF-12-1-0241 and W911NF-15-1-0484. We thank anonymous EMNLP reviewers for constructive feedback. Additionally, we would like to thank Jason Baldbridge for reviewing an early version of this paper, and Kristina Toutanova for helpful discussion.

{
    \bibliographystyle{acl_natbib}
    \bibliography{refs}
}

\clearpage
\appendix
\label{sec:supp}
\section*{Appendix}

\begin{table*}[tbh]
    \centering
    \begin{tabular}{@{\;}l@{\quad}m{6cm}@{\;}m{7.5cm}@{\;}}
    \toprule
    Syntax Tree 
    & 
    \adjustbox{max width=6cm}{
        \Huge
        \Tree [.S \qroof{two dogs}.NP [.VP [.VBP \textcolor{red}{are} ] [.VBG \textcolor{red}{running} ] [.PP [.IN \textcolor{red}{on} ] \qroof{the grass}.NP ] ] ] 
    }
    &
    \adjustbox{max width=7.5cm}{
        \Huge
        \Tree  [.S \qroof{a small pizza}.NP [.VP [.VBN \textcolor{red}{cut} ] [.PP [.IN \textcolor{red}{in} ] [.NN \textcolor{red}{half} ] ] [.IN \textcolor{red}{on} ] \qroof{a white plate}.NP ] ]
    }
    \\
    \midrule
    \addlinespace
    Concept & \texttt{two dogs are running on the grass} & \texttt{a small pizza cut in half on a white plate} \\
    \addlinespace
    Predicate & \texttt{[NP] are running on [NP]} & \texttt{[NP] cut in half on [NP]} \\
    \addlinespace
    Sub-Concepts & \makecell[l]{\texttt{NP1=``two dogs''} \\ \texttt{NP2=``the grass''}} & \makecell[l]{\texttt{NP1=``a small pizza''} \\ \texttt{NP2=``a white plate''}} \\
    \bottomrule
    \end{tabular}
    \caption{Explanatory example of extracting predicates and sub-concepts from a concept }
    \label{tab:example_constituency}
\end{table*}

In the \suppl, we provide details omitted from the main text due to the limited space, including:
\begin{itemize}[topsep=1pt,parsep=0pt,partopsep=4pt,leftmargin=*,itemsep=2pt]
    \item \S~\ref{supp:sec:primitive_crg} describes the implementation details for extracting primitives \& predicates from the constituency tree (\S~2 of the main text).
    \item In \S~\ref{supp:sec:generate_compositional_split}, we describes the details of generating the compositional evaluation splits (\S~5.1 of the main text).
    \item \S~\ref{supp:sec:implementation_details} contains training and architecture details for \ourmethod and baselines (\S~5.1 of the main text).
    \item  \S~\ref{supp:sec:ablation_exps}  includes the ablation studies on learning objectives and margin of MVSA (\S~5.2 of the main text).
\end{itemize}

\section{Extracting Primitives \& Predicates from the Constituency Tree}
\label{supp:sec:primitive_crg}
As mentioned in the main paper, we parse the sentence and convert it into a tree of concepts and primitives. Particularly, we first perform constituency parsing using the self-attention parser~\cite{kitaev2018constituency}. Table~\ref{tab:example_constituency} provides the visualization for two examples of the syntax sub-trees. Next, we perform a tree search (\ie, breadth-first search) on the constituency tree of the current input concept to extract the sub-concepts and predicate functions. Note that this step is applied recursively until we can no longer decompose a concept into any sub-concepts. On a single step of the extraction, we enumerate each node in the constituency tree of current input text expression and examine whether a constituent satisfies the criterion that defines the visually grounded concept.

The concept criterion defined for the Flickr30K and COCO dataset contains several principles: (1) If the constituent is a word, it is a primitive concept if its Part-of-Speech (POS) tag is one of the following: $\{$\texttt{[NN]},\texttt{[NNS]},\texttt{[NNP]},\texttt{[NNPS]}$\}$; (2) If the constituent is a phrase  (with two words or more), it would be a concept when this constituent contains a primitive word (\ie, satisfying condition (1)) and its constituency tag is one of the following: $\{$\texttt{[S]}, \texttt{[SBAR]}, \texttt{[SBARQ]}, \texttt{[SQ]}, \texttt{[SINV]}, \texttt{[NP]}, \texttt{[NX]}$\}$. After all the concepts are extracted, we take the remaining words in the current input text expression as the predicate that combines those concepts and use the tag to represent syntactic blank. Concrete examples can be found in the Table~\ref{tab:example_constituency}. For instance, in the first example, we search the text ``two dogs are running on the grass'' and extract two noun constituents, ``two dogs'' and ``the grass'' as the concepts. We use the remaining text "\texttt{[NP]} is running on \texttt{[NP]}" as the predicate that indicates the semantic meaning of how these two sub-concepts composes into the original sentence.

\section{Details on Generation of Compositional Evaluation Splits}
\label{supp:sec:generate_compositional_split}

As mentioned in the main text, we generate compositional generalization (CG) splits with 1,000 images and 5,000 text queries, maximizing the Compound Divergence (MCD) as \citet{shaw2020compositional}\footnote{We adopt the released code here for the computing compound divergence: https://github.com/google-research/language/tree/master/language/nqg/tasks}, to assess models' capability in generalizing to the data with different predicate distribution. Concretely, we select Flickr30K training data to generate the F30K-MCD split. First, we remove all F30K test data that has unseen primitive concepts to the COCO training data. Next, we collect and count the predicates for each image among all the remaining data over the five associated captions. These predicates correspond to the ``compounds'' defined in ~\cite{keysers2020measuring, shaw2020compositional}, and the objective is to maximize the divergence between compound distribution of the evaluation data to the training data. As a result of this step, we end up with a data set formed with pairs of (image, predicates counts), which are then used for computing the overall compound divergence ($\texttt{CD}_{\textsc{all}}$) to the training dataset. Afterwards, we enumerate over each pair of data, and again compute the compound divergence to the training dataset but with this specific data is removed. We denote the change of compound divergence as $\Delta_i = \texttt{CD}_{i} - \texttt{CD}_{\textsc{all}}$, and use it as an additional score to associate every data. Finally, we sort all the data with regard to the difference of compound divergence $\Delta_i$, and use the top ranking one thousand examples as the maximum compound divergence (MCD) split. The process for generating the COCO-MCD split is symmetrical to the above process, except the data is collected from COCO val+test splits (as it is sufficiently large). Similarly, to generate different CDs for making Figure 4 of the main text, we can also make use of the above data sorted by $\Delta_i$. Concretely, we put a sliding window with 1,000 examples and enumerate over the sorted data to obtain a massive combination of data (we can take a stride to make this computation sparser.) For each window of data, we measure the compound divergence and only take the windows that are at the satisfaction to our criteria. In Figure 4, we keep the windows that has the closest CD values to desired X-axis values for plotting.

\section{Implementation Details of \ourmethod and Baselines}
\label{supp:sec:implementation_details}

\paragraph{Visual feature pre-processing} We follow ViLBERT~\cite{Lu2019ViLBERTPT} that extracts the patch-based ResNet feature using the Bottom-Up Attention model. The image patch feature has a dimension of 2048. A 5-dimension position feature that describes the normalized up-top and bottom-down position is extracted alongside the image patch feature. Therefore, each image region is described by both the image patch feature and the position feature. We extracted features from up to 100 patches in one image.

\paragraph{Text pre-processiong} Following BERT~\cite{Devlin2019BERT}, we tokenize the text using the uncased WordPiece tokenizer. Specifically, we first lowercase the text and use the uncased tokenizer to extract tokens. The tokenizer has a vocabulary size of 30,522. The tokens are then transformed into word embeddings with 768 dimensions. Besides the word embedding, a 768-dimension position embedding is extracted. Both position embedding and word embedding are added together to represent the embedding of tokens.

\paragraph{Training details} We use Adam optimizer~\cite{kingma2014adam} to optimize the parameter of our model. All the models are trained with a mini-batch size of 64. We employ a warm-up training strategy as suggested by ViLBERT~\cite{Lu2019ViLBERTPT}. Specifically, the learning rate is linearly increasing from 0 to $4e-5$ in the first 2 epochs. Then the learning rate decays to $4e-6$ and $4e-7$ after 10 epochs and 15 epochs, respectively. The training stopped at 20 epochs.

\paragraph{Detials of baseline approaches.} The text encoder for both models contains 12 layers of transformers and is initialized from BERT pretrained model using the checkpoint provided by HuggingFace. For ViLBERT, we use the \texttt{[CLS]} embedding from the last layer as text representation $\vy$. We use the average of contextualized text embedding from the last layer as $\vy$ in the VSE model. The visual encoder of VSE contains an MLP model with the residual connection. It transforms the image patch feature into a joint image-text space. The output of the visual encoder is the mean of the transformed image patch features. Unlike VSE, ViLBERT contains 6 layers of transformers for the image encoder and 6 layers of the cross-modal transformer to model the text and image features jointly. We use the embedding of \texttt{[V-CLS]} token from the last layer of the image encoder as the image feature $\vx$. 

\begin{figure}
    \centering
    \includegraphics[width=0.47\textwidth]{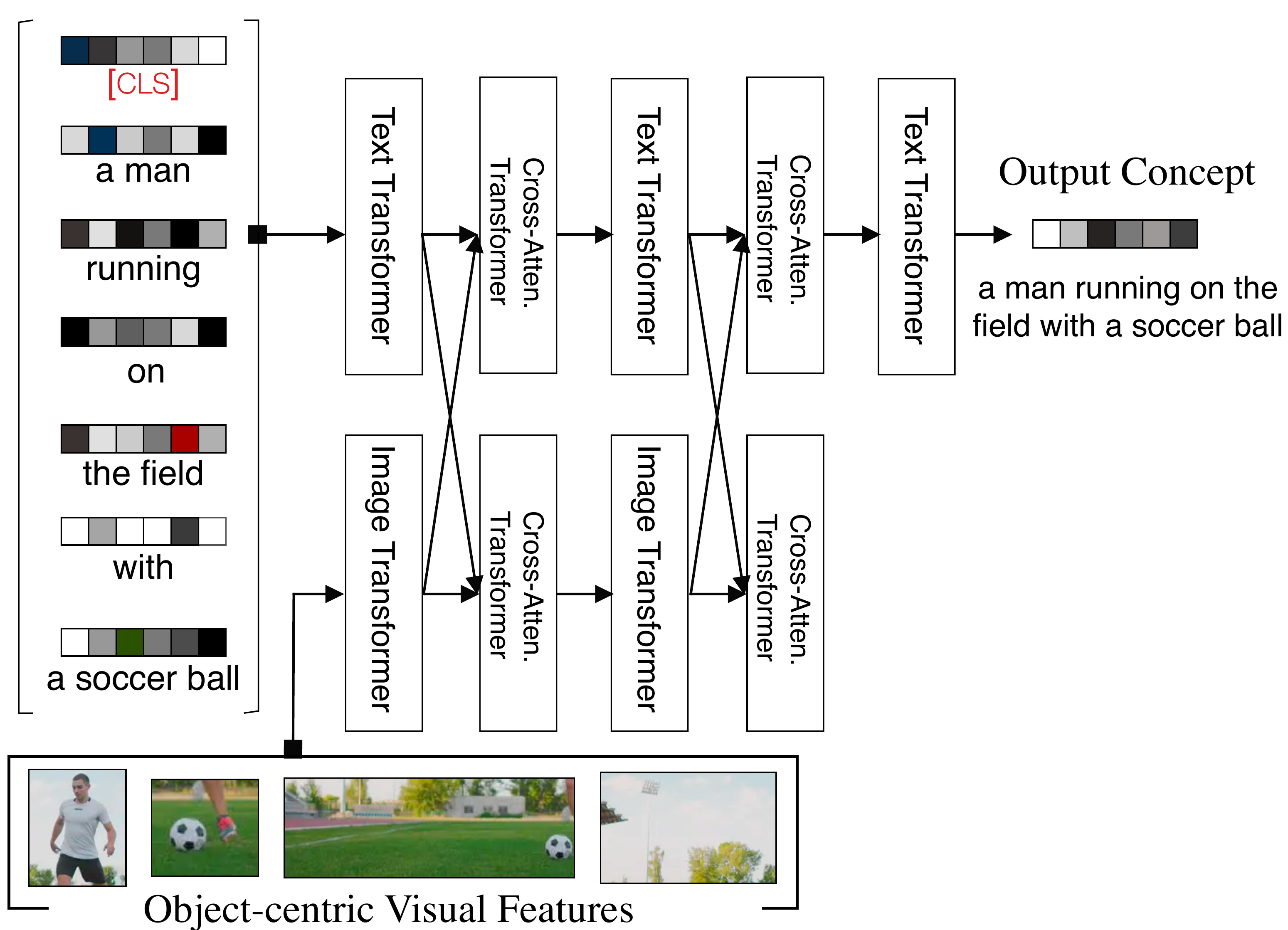}
    \caption{Details of the Composition Transformer model.}
    \label{fig:ct_diagram}
\end{figure}

\paragraph{Details of \ourmethod.} The composer contains four primary learning sub-modules: (1) the \texttt{CrossAtt} model in primitive encoding; (2) the Predicate Transformer (PT) model; (3) the modulator; (4) the Composition Transformer (CT). The details of this sub-modules are list as what follows:

\begin{itemize}[leftmargin=*,topsep=0pt,itemsep=0pt]
    \item \textbf{Primitive encoding.} We implement the \texttt{CrossAtt} model as a one-layer multi-head cross-modal Transformer that contains 768 dimension with 12 attention heads. The query set $Q$ is the sub-word token embeddings of the primitive word, and the key and value set $K$ and $V$ are the union of sub-word token embeddings and the object-centric visual features (which is linearly transformed to have the same dimensionality). We use the average of the contextualized sub-word token embeddings as the final primitive encoding.
    \item \textbf{Predicate Transformer (PT).} We use 6 layers text Transformers with 768 hidden dimension and 12 attention heads to instantiate the Predicate Transformer. This network is initialized with the first 6 layers of a pre-trained BERT model. 
    \item \textbf{Modulator.} We use FiLM~\cite{perez2018film} as the modulator. Specifically, it contains two MLP models with a hidden dimension size of 768 to generate the scale $\va$ and bias vectors $\vb$, using the syntactic placeholders as input. The scale $\va$ and bias $\vb$ are then used to transform the input concept embedding $\vc$ as $\va \odot \vc + \vb$. Here $\odot$ represents the element-wise multiplication. This modulated concept embedding is then projected by another MLP with 768 hidden dimensions, and used for reassembling with the predicate sequence.
    \item \textbf{Composition Transformer (CT).} We follow the architecture of ViLBERT~\cite{Lu2019ViLBERTPT} to design the Composition Transformer (shown in Figure~\ref{fig:ct_diagram}). Specifically, it has interleaved \texttt{SelfAtt} Transformer and \texttt{CrossAtt} Transformer in the network. For example, if we consider a three-layer Composition Transformer, we have a \texttt{SelfAtt} Transformer at the beginning for both modality, followed with a \texttt{CrossAtt} Transformer that interchanges the information between the modality, and then another \texttt{SelfAtt} Transformer that only operates on the text modality. The output embedding of this last text \texttt{SelfAtt} Transformer is then used for computing the visual-semantic alignment scores using the linear regressor $\vtheta$. Thus, when we consider shallower or deeper network, we add or remove the two layers of interleaved \texttt{SelfAtt} and \texttt{CrossAtt} Transformers. The hidden dimension of \texttt{SelfAtt} Transformer is 768, and there is 12 attention heads. 
    The hidden dimension of \texttt{CrossAtt} Transformer is 1024, and there is 8 attention heads.
\end{itemize}

\section{Additional Experiments on \ourmethod}
\label{supp:sec:ablation_exps}

We report additional ablation studies that are omitted in the main paper due to space limitation. In this section, we study {\ourmethod} performance under different MVSA objectives, Negative Log-Likelihood and Hinge loss. Then we study {\ourmethod} performance under different margins of MVSA and Order objectives.


\begin{table}[t]
    \centering
    \small
    {
        \tabcolsep 10pt
	    \begin{tabular}{@{\;\;} ll cccc@{\;\;}}
	        \addlinespace
	        \toprule
	         & & \multicolumn{2}{c}{F30K$\rightarrow$F30K} &
	        \multicolumn{2}{c}{F30K$\rightarrow$COCO} \\
	        \cmidrule(lr){3-4} \cmidrule(lr){5-6}
	        {$\alpha$} & {$\beta$} & {R1} & {R5} & {R1} & {R5} \\ 
	        \midrule
	        \multicolumn{6}{@{}l}{\bf \ourmethod w/ different $\alpha$} \\
	        0.4 & 0.2 & 53.54 & 80.51 & 33.73 & 61.67\\
	        0.6 & 0.2 & 53.44 & 80.21 & 33.89 & 61.05\\
	        0.8 & 0.2 & 54.02 & 80.27 & 33.81 & 63.19 \\
	        \midrule
	        \multicolumn{6}{@{}l}{\bf \ourmethod w/ different $\beta$} \\
	        0.8 & 0 & 53.66 & 80.39 & 33.33 & 61.15 \\
    	    0.8 & 0.2 & 54.02 & 80.27 & 33.81 & 63.19 \\
    	    0.8 & 0.4 & 53.50 & 80.55 & 33.87 & 61.15 \\
	        \bottomrule
	    \end{tabular}
	}
    \caption{Ablation Study on \ourmethod with Different Margin for MVSA and Order Objectives.}
    \label{tab:supp_mvsa_order_margin}
\end{table}

\paragraph{MVSA Objective.} The MVSA objectives can be implemented using NLL loss or Hinge loss. We study the performance of {\ourmethod} under different losses for MVSA in Table~\ref{tab:supp_nll_hinge}. The models are trained with both MVSA and order objectives. We set the margin of order objectives $\beta=0.2$. For the hinge loss, we set the margin $\alpha=0.8$. {\ourmethod} trained with hinge loss in MVSA achieves better performance than the NLL loss in all metrics across both in-domain and cross-dataset generalization settings. Therefore, for all the experiments training with MVSA, we use hinge loss instead.


\begin{table}[t]
    \centering
    \small
    {
        \tabcolsep 10pt
	    \begin{tabular}{@{\;} l cccc@{\;}}
	        \addlinespace
	        \toprule
	         & \multicolumn{2}{c}{F30K$\rightarrow$F30K} &
	        \multicolumn{2}{c}{F30K$\rightarrow$COCO} \\
	        \cmidrule(lr){2-3} \cmidrule(lr){4-5}
	        Loss Function & {R1} & {R5} & {R1} & {R5} \\ 
	        \midrule
	        NLL   & 52.42 & 79.47 & 33.41 & 62.17 \\
	        Hinge Loss & 54.02 & 80.27 & 33.81 & 63.19 \\
	        \bottomrule
	    \end{tabular}
	}
	\caption{Ablation Study MVSA Objective: Comparing NLL to Hinge Loss.}
    \label{tab:supp_nll_hinge}
\end{table}

\paragraph{Ablation study on $\alpha$ and $\beta$.} We study {\ourmethod} performance on the different margin of MVSA and Order objectives. First, we fix the margin of order objectives $\beta$ and tune the margin for MVSA $\alpha$. {\ourmethod} with a larger margin for MVSA achieves better R1 in-domain performance. Alternatively, by fixing the $\alpha$ and tuning $\beta$, {\ourmethod} achieves the best R1 in-domain performance and best R5 in cross-dataset generalization setting with $\beta=0.2$.

\end{document}